\documentclass[journal]{IEEEtran}
\usepackage{hyperref}
\usepackage{amsmath}
\usepackage{color}
\usepackage{array}
\usepackage{stfloats}
\usepackage{amsthm} 
\usepackage{amssymb}
\usepackage{float}
\usepackage{nccmath}
\usepackage{graphicx}
\usepackage{setspace,booktabs}
\usepackage{multicol}
 \usepackage[normalem]{ulem}
\usepackage{multirow}
\usepackage{hyperref}
\usepackage{booktabs}
\usepackage{subcaption}
\usepackage{mwe}    
\usepackage{cite}
\usepackage{caption}
\usepackage{graphicx}
\usepackage[export]{adjustbox}
\usepackage{amsmath,amssymb,amsfonts}
\usepackage{textcomp}
\usepackage{xcolor}
\usepackage[utf8]{inputenc} 
\usepackage{kotex} 

\usepackage{algorithm} 
\usepackage{algpseudocode}
\usepackage{grffile}
\usepackage{balance}

\newcommand{\BfPara}[1]{{\noindent\bf#1.}\xspace}
\begin{document}
\graphicspath{ {./images/} }
\title{Self-Configurable Stabilized Real-Time Detection Learning for Autonomous Driving Applications}

\author{Won Joon Yun,~\IEEEmembership{Graduate Student Member,~IEEE}, Soohyun Park,~\IEEEmembership{Graduate Student Member,~IEEE}, Joongheon Kim,~\IEEEmembership{Senior Member,~IEEE}, and David Mohaisen,~\IEEEmembership{Senior Member,~IEEE}
    \thanks{
    This research is supported by the Institute of Information \& Communications Technology Planning \& Evaluation (IITP) grant funded by the Korea government (MSIT) (2021-0-00467). 
    \textit{(Corresponding authors: Soohyun Park, Joongheon Kim, David Mohaisen)}}
    \thanks{W.J. Yun, S. Park, and J. Kim are with the School of Electrical Engineering, Korea University, Seoul 02841, Korea e-mails: ywjoon95@korea.ac.kr, soohyun828@korea.ac.kr,joongheon@korea.ac.kr.}
    \thanks{D. Mohaisen is with the Department of Computer Science, University of Central Florida, Orlando, FL, USA e-mail: mohaisen@ucf.edu.}
}

\maketitle

\begin{abstract}
Guaranteeing real-time and accurate object detection simultaneously is paramount in autonomous driving environments.
However, the existing object detection neural network systems are characterized by a tradeoff between computation time and accuracy, making it essential to optimize such a tradeoff. 
Fortunately, in many autonomous driving environments, images come in a continuous form, providing an opportunity to use optical flow.
In this paper, we improve the performance of an object detection neural network utilizing optical flow estimation. In addition, we propose a Lyapunov optimization framework for time-average performance maximization subject to stability. It adaptively determines whether to use optical flow to suit the dynamic vehicle environment, thereby ensuring the vehicle's queue stability and the time-average maximum performance simultaneously.
To verify the key ideas, we conduct numerical experiments with various object detection neural networks and optical flow estimation networks. In addition, we demonstrate the self-configurable stabilized detection with YOLOv3-tiny and FlowNet2-S, which are the real-time object detection network and an optical flow estimation network, respectively. 
In the demonstration, our proposed framework improves the accuracy by 3.02\%, the number of detected objects by 59.6\%, and the queue stability for computing capabilities.
\end{abstract}


\section{Introduction}
Real-time object detection algorithms are getting a lot of attention in academia and industry for the broad set of applications that employ them, including autonomous driving and factory automation~\cite{its1,its5,its6,its7,its8}.
Among those algorithms, the multi-object detection algorithm is widely used. The fundamental ideas of the algorithm mentioned above consist of two procedures; (i) \textit{bounding box} and \textit{confidence} are used for determining whether objects exist or not; then (ii) the concept of \textit{class probability map} is used for the classification of the detected objects~\cite{yolo}.
The multi-object detection algorithms are based on convolutional neural networks (CNN). Nowadays, sophisticated algorithms are proposed to improve the accuracy of the network, such as YOLOv4 and DyHead~\cite{ YOLOv4, DyHead}. However, there is a tradeoff between inference time and accuracy. The object detection algorithm with many convolution layers performs better than others while increasing the computation time. 
In contrast, the simple network that requires less computation is suitable for real-time object detection applications while sacrificing specific detection accuracy.

For autonomous driving applications, the major learning-based system research topic, it is necessary to ensure the system can use and relate to the driving environment.  The driver can record the environment in a real-time video composed of continuous image arrivals.
To use the continuous image arrivals, \textit{optical flow} is one of the best ways to model and process sequential continuous image arrivals~\cite{opticalflow}. There are two types of optical flow; (i) One is a sparse optical flow that presents a partial motion of pixels. (ii) The other is a dense optical flow that indicates the full motion of pixels. Dense optical flow has higher accuracy than sparse optical flow, while the speed is lower. In the recent decade, the neural network application on calculating dense optical flow (\textit{e.g.}, FlowNet) emerges to obtain a dense flow map with guarantying high accuracy and low speed~\cite{flownet}.
This network is a CNN-based neural network that receives two consecutive images of a video as an input and returns the information of pixel displacement as an output. The optical flow estimation networks (OFEN), similar to object detection, have a tradeoff between the computation time (\textit{i.e.}, delay) and accuracy.

\begin{figure}[t!]
\centering
\setlength{\tabcolsep}{2pt}
\renewcommand{\arraystretch}{0.2}
\begin{tabular}{cc}
\includegraphics[page=1, width=0.45\linewidth]{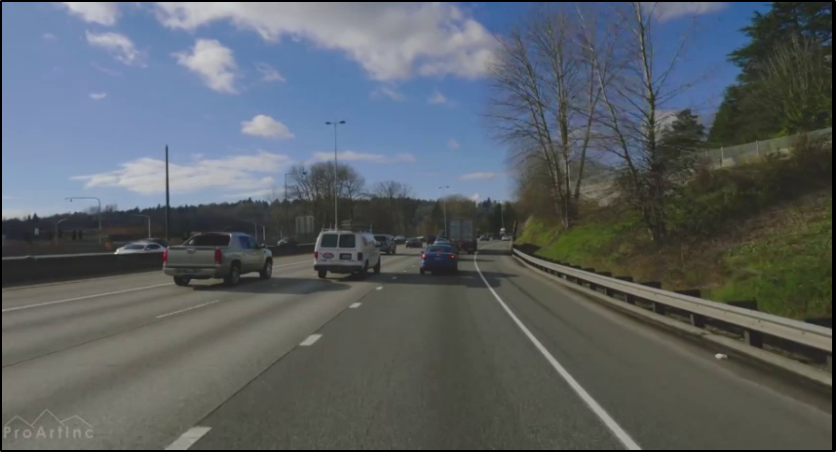} &
\includegraphics[page=1, width=0.45\linewidth]{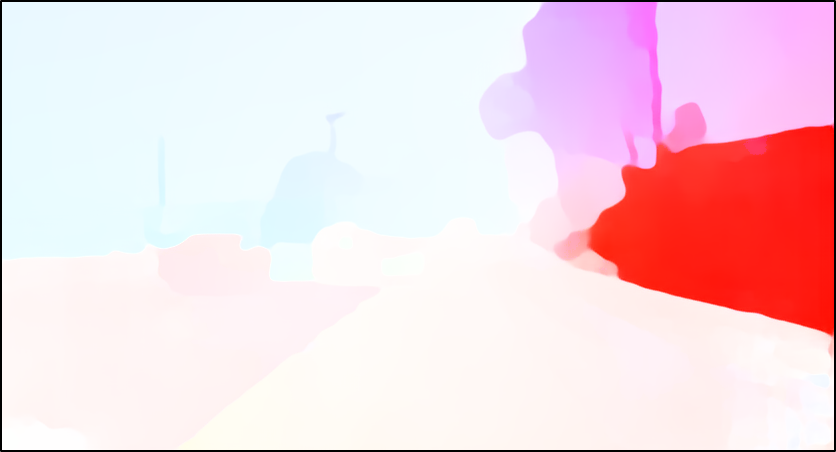} 
\tabularnewline
\tabularnewline
\footnotesize (a) Input image & \footnotesize (b) Dense optical flow field
\tabularnewline
\tabularnewline
\includegraphics[page=1, width=0.45\linewidth]{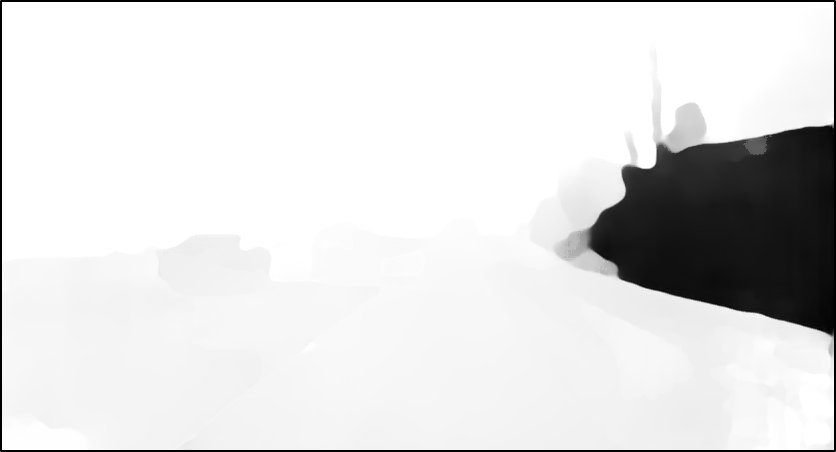} &
\includegraphics[page=1, width=0.45\linewidth]{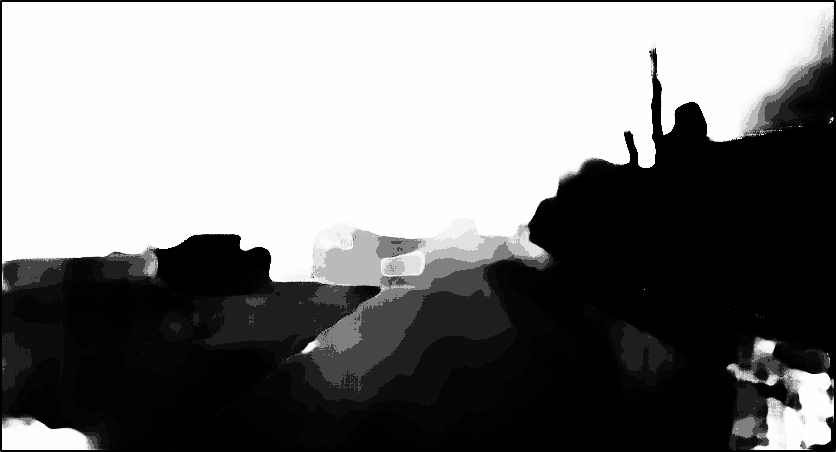}
\tabularnewline
\tabularnewline
\footnotesize (c) Flow map & \footnotesize (d) The processed flow map \textbf{(Ours)}
\end{tabular}
\footnotesize \caption{A snapshot of optical flow in an actual driving environment.}
\label{fig:flo_description}
\end{figure}

Motivated by the \textit{``continuity"} property of videos in the object detection research domain, this paper investigates object detection and optical flow consolidation for the driving environment. As shown in Fig.~\ref{fig:flo_description}(a)--(c), we empirically find that leveraging optical flow can be a solution for improving the existing object detection networks (ODN). However, applying an optical flow on object detection directly is challenging due to following reasons.
First, the driving environment is generally uncertain, \textit{e.g.}, the vehicle can be unexpectedly driving or stopping.
Second, the driver state is time-varying. If the observer is static, the moving object is detected by optical flow. On the other hand, and according to Fig. \ref{fig:flo_description}(c), \textit{i.e.}, if the observer is moving, the moving objects may appear static when moving in the same direction.
Additionally, we consider the computing capacities corresponding to object detection in the driving environment. The received images are time-varying. Thus, the number of objects in the changes every moment. However, the number of objects is proportional to the inference time of an object detection network~\cite{num_object}. 

In light of the issues above, we propose a novel real-time object detection model called \textit{Hybrid} by leveraging optical flow. Then, we design a self-configurable stabilized framework with Lyapunov optimization considering computational overheads \cite{book2010sno}. Our framework aims to guarantee time-average object detection performance maximization by deciding whether to use optical flow considering computation time (delay) based on the driving environment.

\BfPara{Contributions} The key contributions of the proposed algorithm in this paper are as follows. 
\begin{itemize}
    \item We first propose a fusion of object detection and optical flow. We further suggest a novel flow map processing algorithm (see \textbf{Algorithm~\ref{alg:confidencemask}}), which is suitable for a time-varying driving environment. 
    \item We propose a novel self-configurable framework for autonomous driving. Our architecture provides a time-average sequential optimal decision-making under the tradeoff between performance and delay. Furthermore, one of the main advantages of the Lyapunov optimization-based algorithm is low-complexity operation (see \textbf{Algorithm~\ref{alg:dpp2}}). Thus, our proposed algorithm is suitable for real-time computation in a fast-moving autonomous driving environment. 
\end{itemize}

\BfPara{Organization} The rest of this paper is organized as follows. 
Sec.~\ref{sec:2} proposes a stabilized real-time object detection adaptation algorithm for autonomous driving.
Sec.~\ref{sec:3} evaluates the performance of the proposed algorithm.
Sec.~\ref{sec:4} concludes this paper and presents future research directions.
The notations used in this paper are listed in Table~\ref{tab:notation}.

\begin{table}[t!]
    \caption{List of Notations}
    \label{tab:notation}
    \centering
    \footnotesize
    \begin{tabular}{c|l}
        \toprule[1pt]
        \textbf{Symbol} & \textbf{Description}\\\midrule[1pt]
        $K$ & The number of bounding boxes.\\
        $\mathbf{F}_{M,N}$ & $M \times N$-sized flow map.\\ 
        $\mathbf{B}_{M,N,K}$ &  $K$ bounding boxes of size $M \times N$.\\ 
        $Q$ & A queue-backlog.\\ 
        $a[t]$ & An arrival process at time $t$.\\ 
        $H$ & A hybrid model leveraging optical flow.\\
        $T$ & An object detection model.\\
        $\alpha[t]$ & A detection model, $\forall \alpha[t] \in \mathcal{A} \equiv \{H,T\}$.\\
        $b(\alpha[t])$ & A service process with $\alpha[t]$ at time $t$.\\ 
        $P(\alpha[t])$ & The object detection accuracy.\\
        \bottomrule[1pt]
     \end{tabular}
\end{table}

\begin{figure*}[t!]
\includegraphics[width=1.0\linewidth]{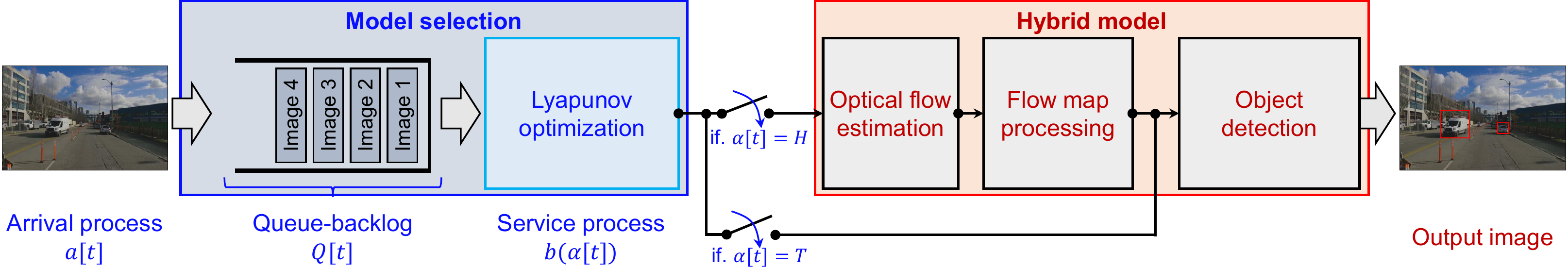}
\caption{The system model consists of two components, \textit{i.e.}, (1) \textit{Hybrid model}, which utilizes optical flow estimation in object detection model, and (2) \textit{Model selection} which is based on Lyapunov optimization. Our proposed method determines whether the model uses optical flow estimation and flow map processing or not.}
\label{fig:systemmodel}
\end{figure*}

\section{Stabilized Real-Time Object Detection for Autonomous Driving Applications}\label{sec:2}
This section introduces our proposed stabilized real-time object detection for autonomous driving, consisting of two parts, \textit{i.e.,} the Hybrid model in Sec.~\ref{sec:2-1} and model selection in Sec.~\ref{sec:2-2}. Fig.~\ref{fig:systemmodel} briefly illustrates our proposed framework.

\subsection{Hybrid Model}\label{sec:2-1}
\subsubsection{Object Detection \& Optical Flow Estimation in the Nutshell}
We present the object detection and optical flow estimation in the nutshell. ODN takes an image as an input and returns bounding boxes $\mathbf{B}_{M,N,K}$ where the component represents confidence score $c_{i,j,k} \in [0, 1]$: $\forall i \in \mathbb{N}[1,M]$, $\forall j \in \mathbb{N}[1,N]$, $\forall k \in \mathbb{N}[1,K]$. Among all bounding boxes, the highest scored bounding box of which the confidence score exceeds the confidence threshold, is regarded as the object detected~\cite{yolo}.

As shown in Fig.~\ref{fig:flo_description}(a), the dense optical flow field is obtained by calculating pixel displacement of two consecutive images via OFEN. As shown in Fig.~\ref{fig:flo_description}(b), The flow map $\mathbf{F}_{M, N}$ indicates the magnitude of dense optical flow field, which is a matrix with the size of $ M \times N $ written as follows:
\begin{equation}
   \mathbf{F}_{M,N} = \begin{pmatrix}
e_{1,1} & e_{1,2} & \cdots & e_{1,N} \\
\vdots  & \vdots  & \ddots & \vdots  \\
e_{M,1} & e_{M,2} & \cdots & e_{M,N} 
\end{pmatrix},
\end{equation} 
where $e_{i,j}$ stands for the magnitude of pixel motion of position $(i,j)$. Note that each element $e_{i,j}$ has a real value where $\forall e_{i,j} \in (-\infty, + \infty)$: $\forall i \in \mathbb{N}[1, M]$ and $\forall j \in \mathbb{N}[1, N]$.
\subsubsection{Observation from Optical Flow}
According to Fig.~\ref{fig:flo_description}(c), the flow map has a linear increase/decrease pattern in each object (\textit{e.g.,} car or truck) and its background. 
Our initial insight is that if flow map and confidence are used as an additional condition for determining the existence of an object, we will be able to perform better than the \textit{de facto} ODN. Suppose that the confidence value of ODN is small, but there is an object detected in the flow map. If so, lowering the confidence threshold for the cell makes it possible to detect objects that ODN cannot detect. To enhance the performance of ODN, we design \textit{Hybrid}, which is the combination of optical flow and object detection. We elaborate on utilizing flow map into object detection next.

\subsubsection{Flow Map Processing} 
This subsection introduces the flow map processing algorithm to apply the Hybrid to the road driving environment.
All elements of $\mathbf{F}_{M,N}$ are in $(-\infty, + \infty)$, thus $\mathbf{F}_{M,N}$ should be min-max normalized.
\begin{equation}
\label{eq:normalize}
    \mathbf{F}_{M,N} \leftarrow \{ \mathbf{F}_{M,N} - e_{\min}\cdot\textbf{1}_{M,N} \},
\end{equation}
where $\textbf{1}_{M,N}$ stands for $M\times N$-sized matrix of ones, and $e_{\min}$ denotes the minimum value of $\forall e_{i,j}$.
The closer the values at both ends, \textit{i.e.}, $e_{\max}, e_{\min}$, the more dynamic pixel information is present, and the central value is static pixel information. As $e_{i,j}$ approaches the maximum value of $\mathbf{F}_{M,N}$, $e_{\max}$, or the minimum of $\mathbf{F}_{M,N}$, $e_{\min}$, dynamic pixel information exists. As it approaches the median value of $\mathbf{F}_{M,N}$, $e_{\mathrm{median}}$, static pixel information exists. This behavior is expressed as follows: 
\begin{equation}
\label{eq:normalize2}
    \mathbf{F}_{M,N} \leftarrow |\mathbf{F}_{M,N} - e_{\mathrm{median}} \cdot\textbf{1}_{M,N}  |
\end{equation}
After this procedure, the information of the moving pixel comes out closer to $e_{\max}$, and the information of the static pixel comes out closer to $e_{\min}$.
Since the purpose of this algorithm is to consider only the moving pixel information, we present the process of removing the static pixel information;
\begin{eqnarray}
\mathbf{F}_{M,N} &\leftarrow& \frac{1}{1+\mathrm{exp}(-\mathbf{F}_{M,N})}.  \label{eq:classify}
\end{eqnarray} 
Then, flatten $\mathbf{F}_{M,N}$ to $\mathbf{f}_{MN}$, and duplicate $\mathbf{f_{MN}}$ as much as the number of bounding boxes $K$. Eventually, the vectorized flow map denoted as $\mathbf{f}_{MNK}$, is applied to the confidence threshold as follows:
\begin{equation}
\label{eq:vectorized_cth}
    \textbf{c}_{th} = \frac{c_{th} }{1+\mathrm{exp}\left(2\cdot  \mathbf{f}_{MNK} \right)}
\end{equation}
where $\textbf{c}_{th}$, and $c_{th}$ stand for the vectorized confidence threshold, and the scalar confidence threshold, respectively.
Finally, leveraging $\textbf{c}_{th}$ as a criterion for object detection, ODN utilizes optical flow.

\begin{algorithm}[t!]
\caption{Flow Map Processing}
\label{alg:confidencemask}
\small
\begin{algorithmic}[1]
\Statex $\hspace{-1.5em}\textbf{Input:}$ $\mathbf{F}_{M,N}$ // $M \times N$ flow map matrix
\Statex $\hspace{-1.5em}\textbf{Output:}$ $\textbf{c}_{th}$ // ${M \times N \times K}$-sized confidence threshold vector
\For{$\forall e_{i,j}$ in $\mathbf{F}_{M,N}$}
\State $e_{i,j} \leftarrow 
e_{i,j}-\min(\mathbf{F}_{M,N})$; 
\EndFor
\For{$\forall e_{i,j}$ in $\mathbf{F}_{M,N}$}
\State $e_{i,j} \leftarrow 
|e_{i,j}-\textsf{median}(\mathbf{F}_{M,N})|$; 
\EndFor
\For{$\forall e_{i,j}$ in $\mathbf{F}_{M,N}$}
\State $e_{i,j} \leftarrow \textsf{sigmoid}(e_{i,j})$; // Equation 4
\EndFor
\State Flatten $\mathbf{F}_{M,N}$ into vector $\mathbf{f}_{MN}$;
\State Replicate $\mathbf{f}_{MN}$ $K$ times, then make $\mathbf{f}_{MNK}$;
\State $\textbf{c}_{th} \leftarrow \frac{c_{th}}{1+\mathrm{exp} (2 \cdot \mathbf{f}_{MNK})}$;
\end{algorithmic}
\end{algorithm}

The pseudo-code of the proposed flow map processing algorithm is presented in \textbf{Algorithm~\ref{alg:confidencemask}}. From (line 1) to (line 3), the elements of flow map $\mathbf{F}_{M,N}$ are normalized. From (line 4) to (line 9), the component of the static object in $\mathbf{F}_{M,N}$ is diminished. In (line 10), $\mathbf{F}_{M,N}$ is scaled to $\mathbf{F}_{S,S}$ using the bicubic interpolation~\cite{bicubic}. From (line 11) to (line 12), the processed flow is converted to the same size as the output size of YOLOv3-tiny. From (line 13) to (line 14), the process of making vectorized confidence threshold $\textbf{c}_{th}$ is expressed using \eqref{eq:vectorized_cth}. The computational complexity of \textbf{Algorithm~\ref{alg:confidencemask}} is $O(MN)$ for flow map with the size $M\times N$.

\subsection{Lyapunov Optimization-based Model Selection}\label{sec:2-2}
In the driving environment, a driving state and a stationary state exist. If the optical flow is used while driving, a Hybrid can improve the performance. However, when the optical flow is used in the stationary state, the pixel displacement does not exist, and the time to calculate it is wasted. Also, in the case of many objects, for example, there is the possibility of being pushed out in real-time suitability because the computation amount increases proportionally. In other words, Hybrid and ODN have a tradeoff between the computation time (\textit{i.e.}, delay) and object detection accuracy. 
Therefore, the Lyapunov optimization framework is designed to increase the system's stability by observing the queues and performance of the two networks and making the right decision.
\subsubsection{Lyapunov Optimization Framework}
This section introduces our proposed Lyapunov optimization framework, aiming at time-average detection performance maximization subject to system stability.
We can design a time-average optimization framework considering stability by stabilizing the drift. In this case, and according to the Lyapunov optimization framework, the delay can be modeled by queue, where Lyapunov drifts can then again model the queue dynamics. By observing the queue-backlog and performance of every frame, the framework can use the Lyapunov optimization to select the next step deep learning object detection model, which is a sequential time-average optimal decision-making.
The queue dynamics in the system $Q[t]$ are characterized as follows: 
\begin{equation}
    Q[t+1] \triangleq \max\{Q[t]+a[t]-b(\alpha[t]),0\}, \label{eq:queue}
\end{equation}
where 
    $Q[t]$ is a queue-backlog size at time $t$ where $Q[0] = 0$ and
    $a[t]$ is an arrival process at $Q[t]$ at $t$.  
    This arrival process is the received video streams in the system (i.i.d. random events).

In \eqref{eq:queue}, $b(\alpha[t])$ is a service process at $Q[t]$ when our model selection decision is $\alpha[t]$ at $t$. 
With the Hybrid, the processing of $Q[t]$ will be relatively smaller compared to the case where ODN is used.
The computational cost for ODN and Hybrid is used as the weights, as follows:
\begin{equation}
b(\alpha [t])=
\begin{cases}
w_1 , & \mbox{if }\alpha[t]=H \\ 
w_2 , & \mbox{if }\alpha[t]=T    
\end{cases}, 
\end{equation}
where $H$ stands for the Hybrid and $T$ stands for ODN, respectively.
Here, $a[t]$ is modeled as the ratio of fps per cycle to default fps: 
\begin{equation}
    a[t] = w_{fps} \cdot p[t] , 
\end{equation}
where $p[t]$ and $w_{fps}$ stand for the time per cycle of the network and default fps, respectively.
 In addition, the object detection accuracy is defined based on the total number of objects detected, the number found correctly, the number found incorrectly, the number of overlapping objects, and the ratio of truly detected~\cite{performance}. Thus, the performance of the detection model is at time-step $t$ is modeled as follows: 
\begin{equation}
P(\alpha [t])=
\begin{cases}
w_p \cdot \textsf{num}_H(\textsf{object}), & \mbox{if }\alpha[t]=H \\ 
\textsf{num}_T(\textsf{object}), & \mbox{if }\alpha[t]=T
\end{cases}, \label{eq:exp2}
\end{equation}
where $w_p$, and $\textsf{num}_{(\cdot)}(\textsf{object})$ stand for the detection accuracy ratio of $H$ and $T$, and the number of detected object, respectively.

In our reference system model, the model that processes video streams fast, whereas the object detection accuracy is relatively low, should be used if $Q[t]$ is near overflow. 
On the other hand, the model which processes video streams with high object detection accuracy while taking more time should be used if $Q[t]$ is near zero. 
Eventually, we can observe the tradeoff between our objective (\textit{i.e.}, object detection accuracy) and stability. 
This paper designs an object detection model selection algorithm for the time-average detection accuracy maximization while guaranteeing queue stability.
The mathematical program for maximizing the time-average object detection accuracy, \textit{i.e.}, $P(\alpha[t])$, is as follows:
\begin{eqnarray}
\max: & & \lim_{t\rightarrow\infty}\sum_{\tau=0}^{t-1} P(\alpha[\tau]), 
\label{eq:opt}
\\
\text{subject to} & & \lim_{t\rightarrow\infty}\frac{1}{t}\sum_{\tau=0}^{t-1} Q[\tau]<\infty \text{ (queue stability)}.
\end{eqnarray}
According to this tradeoff, the Lyapunov optimization theory-based drift-plus-penalty (DPP) algorithm~\cite{tvt2019minseok,ton2016joongheon,tmc2019jonghoe} maximizes the time-average utility subject to queue stability.
Here, the Lyapunov function is defined as 
\begin{equation}
L(Q[t]) \triangleq \frac{1}{2}Q^2[t]
\end{equation}
and $\Delta(.)$, the conditional quadratic Lyapunov function, is defined as  \begin{equation}
\mathbb{E}[L(Q[t+1])-L(Q[t])| Q[t]]
\end{equation}
also called the drift on $t$. 
According to \cite{book2010sno}, this dynamic policy is designed to achieve queue stability by minimizing an upper bound on DPP (\textit{i.e.}, minimizing the negative value of $P(\alpha[t])$), which is given by
\begin{equation}
\Delta(Q[t]) + V \mathbb{E} \Big[ -P(\alpha[t]) \Big],
\end{equation}
where $V$ is a tradeoff coefficient. 
The upper bound on the drift of the Lyapunov function at $t$ is derived as follows:
\begin{align}
    &L(Q[t+1]) - L(Q[t]) = \frac{1}{2}\Big( Q([t+1]^2 - Q[t]^2 \Big) \\
    &~\leq \frac{1}{2} \Big( a[t]^2 + b(\alpha[t])^2 \Big) + Q[t] (a[t] - b(\alpha[t])).
\end{align}
Therefore, the upper bound of the conditional Lyapunov drift can be derived as follows:
\begin{align}
    \Delta(Q(t)) &= \mathbb{E}[L(Q[t+1]) - L(Q[t]) | Q[t]] \nonumber \\
    &\leq C + \mathbb{E}\Big[ Q[t](a[t] - b(\alpha[t]) \Big| Q[t] \Big],
\end{align}
where $C$ is a constant given by
\begin{equation}
    \frac{1}{2}\mathbb{E}\Big[ a[t]^2 + b(\alpha[t])^2 \Big| Q[t] \Big] \leq C,
\end{equation}
which assumes that the arrival and departure process rates are upper bounded.
Due to the fact that $C$ is a constant and the arrival process $a[t]$ is not controllable, minimizing the upper bound on DPP becomes
\begin{equation}
    V \mathbb{E}\Big[  -P(\alpha[t]) \Big] - \mathbb{E}\Big[ Q[t]\cdot b(\alpha[t]) \Big]. 
\end{equation}
Thus, the time-average maximization problem can be re-formulated as
\begin{equation}
    V \mathbb{E}\Big[ P(\alpha[t]) \Big] + \mathbb{E}\Big[ Q[t]\cdot b(\alpha[t]) \Big]. 
    \end{equation}

\begin{algorithm}[t]
 \caption{Stabilized Detection Accuracy Maximization}
\label{alg:dpp2}
\small
\begin{algorithmic}[1]
\Statex $\hspace{-1.5em}\textbf{Initialize:}$ $t\leftarrow 0$; $Q[t]\leftarrow 0$;
\Statex $\hspace{-1.5em}\textbf{Decision Action:}$ $\forall \alpha[t]\in\mathcal{A}\equiv\{H,T\}$ 
\Statex $\hspace{-1.5em}\textbf{Stabilized Object Detection Accuracy Maximization:}$
\While{$t\leq T$} // $T$: operation time
    \State Observe $Q[t]$;
    \State $\mathcal{T}^{*} \leftarrow \infty$;
    \For{$\alpha[t]\in \mathcal{A}$}
        \State
        $\mathcal{T}
            \leftarrow V\cdot P(\alpha[t]) + Q[t]b(\alpha[t])$;
        \If {$\mathcal{T} \leq \mathcal{T}^{*}$}
            \State            $\mathcal{T}^{*}\leftarrow\mathcal{T}$; \State  $\alpha^{*}[t]\leftarrow \alpha[t]$;
        \EndIf
    \EndFor
\EndWhile
\end{algorithmic}
\end{algorithm}

Here, the concept of the maximization of the expectation is used; therefore, this result can be maximized by an algorithm that observes the current queue state $Q[t]$ and determines $\alpha[t]$ at every slot $t$, as follows:

\begin{equation}
\boxed{
    \alpha^{*}[t+1]\leftarrow
    \arg\max_{\alpha[t]\in\mathcal{A}}
    \left[
        V\cdot P(\alpha[t]) + Q[t]b(\alpha[t])
    \right]}
\label{eq:lyapunov-final}
\end{equation}
where $\mathcal{A}\equiv\{H,T\}$ is the set of all possible object detection models, $\alpha^{*}[t]$ is the optimal object detection model selection decision at $t$, and $V$ is the tradeoff coefficient between the processing accuracy and queue stability.
To verify whether the \eqref{eq:lyapunov-final} works correctly or not, we provide the following two cases.
\begin{itemize}
    \item \textit{Case 1:} Suppose that $Q[t]\approx \infty$. Then, \eqref{eq:lyapunov-final} tries to maximize $b(\alpha[t])$, thus the processing should be accelerated for satisfying the queue stability, and the object detection model at $t$ ($\alpha[t]$) is selected, which is the fastest one.
    \item \textit{Case 2:} Suppose that $Q[t] =0$. Then, \eqref{eq:lyapunov-final} tries to maximize $P(\alpha[t])$, thus the algorithm pursues the performance accuracy improvements, and the object detection model at $t$ ($\alpha[t]$) is selected, which is the most accurate one. 
\end{itemize}


\subsubsection{Pseudo-Code and Complexity}\label{sec:2-2B}
The pseudo-code of the proposed object detection model selection algorithm is presented in \textbf{Algorithm~\ref{alg:dpp2}}. 
All variables and parameters are initialized from (line 1) to (line 3). The algorithm works in each unit time as shown in (line 4). In (line 5), the current queue-backlog $Q[t]$ is observed to be used in \eqref{eq:lyapunov-final}. From (line 7) to (line 12), the main computation procedure for \eqref{eq:lyapunov-final} is described. 
Because our proposed algorithm solves a closed-form equation with the number of decision actions, \textit{i.e.}, the number of elements in $\mathcal{A}$, the run-time computational complexity is only $O(N)$. Thus, it is clear that our algorithm guarantees a low computational complexity.

\section{Performance Evaluation}\label{sec:3}
This section presents the implementation of Hybrid (refer to Sec.~\ref{sec:3-1}) and the performance of the proposed model selection (refer to Sec.~\ref{sec:3-2}), respectively. Note that the experiment settings are presented in Table~\ref{tab:settings}. Due to the real-time issue, we adopt YOLOv3-tiny and FlowNet2-S as the real-time ODN and OFEN, respectively. The demo video is available in~\cite{youtube}.
\subsection{Implementation of Hybrid}\label{sec:3-1}

\begin{figure}[t!]
\centering
\setlength{\tabcolsep}{2pt}
\renewcommand{\arraystretch}{0.2}
\centering \includegraphics[page=1, width=\columnwidth]{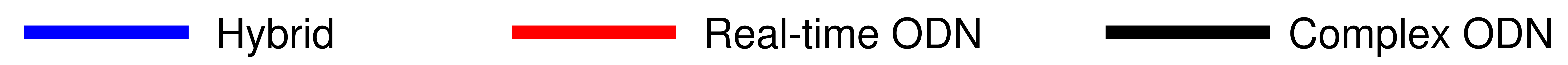}
\begin{tabular}{cc}
    \includegraphics[page=1, width=0.5\columnwidth]{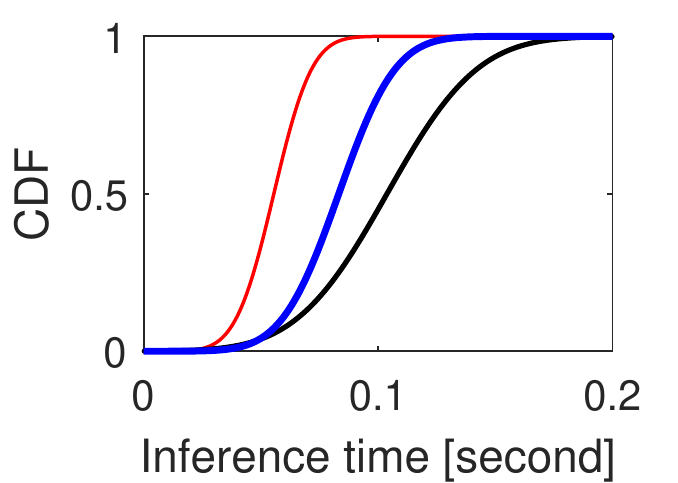} & \includegraphics[page=1, width=0.5\columnwidth]{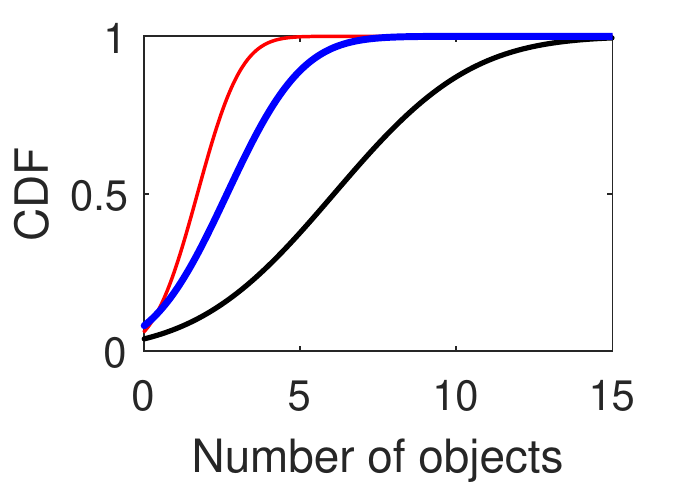} \\
    \\
    \\
\small (a) CDF of the inference time & \small (b) CDF of the number of objects\\
\end{tabular}
\caption{The three models' cumulative density functions ($y$-axis values) depend on the inference time and the number of objects in the GPU setting.}
\label{fig:performance}
\end{figure}

\begin{table}[t!]
\footnotesize
    \centering
        \caption{Experiment setting.} 
    \label{tab:settings}
    \begin{tabular}{c|r}
    \toprule[1pt]
    \textbf{Specitications} & \textbf{Settings}\\\midrule
    \multicolumn{1}{l|}{Dataset} & nuScenes-mini~\cite{nuscenes}, 4K Driving~\cite{dataset} \\\midrule
    \multicolumn{1}{l|}{OS}  & Windows 10 Pro \\\midrule
    \multicolumn{1}{l|}{Processor (CPU \& GPU)} & Intel Xeon E5-2638, Nvidia RTX 2080Ti\\\midrule
    \multicolumn{1}{l|}{Dev. environment} & Python 3.6, Pytorch 1.4, OpenCV 4.2\\\midrule
    \multicolumn{1}{l|}{Object detection networks}  & YOLOv3/v3-tiny/v4/v7 pretrained \\
    & with COCO data\\\midrule
    \multicolumn{1}{l|}{Optical flow } & FlowNet2/2C/2S pretrained\\ 
    \multicolumn{1}{l|}{estimation networks} & with KITTI data\\ \midrule
    \multicolumn{1}{l|}{Lyapunov coefficient ($V$)} & 90\\ \midrule
    \multicolumn{1}{l|}{Confidence threshold ($c_{th}$)}  & 0.5 \\ \midrule
    \multicolumn{1}{l|}{NMS threshold}  & 0.2 \\ \midrule
    \multicolumn{1}{l|}{Weights ($w_1, w_2, w_{fps},w_p$)} & (3.64, 2.41, 30.0, 1.005)\\ 
    \bottomrule[1pt]
    \end{tabular}
\end{table}
\begin{table}[t!]
\footnotesize
    \centering
        \caption{Performance of Hybrid with various models and nuScenes-mini dataset (mAP50).}
\label{tab:performance-general}
\begin{tabular}{@{}l|cccc@{}}
\toprule[1pt]
\centering
\hspace{1em}\textbf{ODN/OFEN} & w/o. OFEN & FlowNet2S & FlowNet2C & FlowNet2\\\midrule
    \multicolumn{1}{l|}{YOLOv3-tiny~\cite{yolov3}}  & 2.40 & 2.45 & 2.42 & 2.47 \\
    \multicolumn{1}{l|}{YOLOv3~\cite{yolov3}}  & 13.65 & 13.66 & 13.66 & 14.20  \\
    \multicolumn{1}{l|}{YOLOv4~\cite{YOLOv4}}  & 15.62 & 15.65 & 15.66 & 16.07 \\ 
    \multicolumn{1}{l|}{YOLOv7~\cite{YOLOv7}}  & 14.45 & 14.48 & 14.47 & 14.54 \\ 
    \bottomrule[1pt]
    \end{tabular}
\end{table}
We propose three main experiments to verify the performance of Hybrid. First, we investigate the impact of flow map on various ODNs with a public driving dataset (\textit{i.e.}, nuScenes-mini \cite{nuscenes}). Second, we investigate the performance of Hybrid with high-resolution driving dataset (\textit{i.e.}, Youtube 4K driving~\cite{dataset}) by comparing the performance of the real-time ODN (\textit{i.e.,} YOLOv3-tiny) and more complex ODN (\textit{i.e.,} YOLOv3). Finally, we conduct an ablation study of optical flow estimation. We measure the time taken for one cycle and the number of detected objects per image for performance evaluation.\\
\BfPara{Impact of flow map on detector networks} We conduct the performance evaluation to investigate the overall performance. In this paper, we adopt nuScenes-mini dataset for evaluation with various real-time ODN models, \textit{e.g.}, YOLOv3, YOLOv4, YOLOv5, and YOLOv7. In addition, we use various optical flow estimation networks, \textit{e.g.}, FlowNet2, FlowNet2C, and FlowNet2S. All ODN networks and OFEN networks are pretrained with COCO dataset and KITTI dataset, respectively. The overall performance is presented in Tab.~\ref{tab:performance-general}. As shown in Tab.~\ref{tab:performance-general}, all OFEN improve ODN networks. Especially, Hybrid composed of YOLOv3 and FlowNet2 achieves 4.02\% performance gain (\textit{i.e.}, 0.55\% mAP50 score) compared to YOLOv3 only. The state-of-the-art YOLOv7 shows the least performance gain 0.6\% (\textit{i.e.}, 0.09\% mAP50 score). In addition, more complex OFEN (\textit{i.e.}, FlowNet2) shows the highest performance gain, the lightest OFEN (\textit{i.e.}, FlowNet2S) shows the lowest performance gain. In summary, our flow map processing algorithm enhances the performance of various ODNs in the driving environment without training/fine-tuning ODNs nor OFENs.\\
\BfPara{Feasibility study of Hybrid} To figure out the feasibility of Hybrid concerning object detection, we test three models (\textit{i.e.}, Hybrid, YOLOv3-tiny, and YOLOv3) with a road driving dataset which consists of 27k frames~\cite{dataset}. We measure the inference time and number of objects per image. 
 Fig.~\ref{fig:performance} shows the result in the GPU setting. We find a tradeoff between inference time (\textit{i.e.,} delay) and the number of the detected object (\textit{i.e.,} performance). In addition, the total number of detected objects is 46.3k for real-time ODN, 72.9k for Hybrid, and 164k for complex ODN, respectively. On average, the inference time in both GPU and CPU settings is high in the order of complex ODN, Hybrid, and real-time ODN as shown in Table~\ref{tab:inf_time}. All three models satisfy real-time in the GPU setting, whereas only the complex ODN does not in the CPU setting. In summary, the Hybrid significantly outperforms two comparison models in the mobile platforms where GPU is incapable.\\
\BfPara{Ablation study of optical flow estimation} To verify the accuracy of the Hybrid model, we design the experiment with humans observing and recording the results of two models (Hybrid and ODN). The road driving dataset, consisting of 2k frames, is used in the corresponding experiment~\cite{dataset}. We set the performance evaluation criteria for the total number of objects detected, the number found correctly, the number found incorrectly, the number of overlapping objects, and the ratio of truly detected. The only difference is whether the simulation leverages optical flow estimation. Tab.~\ref{tab:my_label} shows the performance evaluation results of the ablation study. The total number of detected objects is $1.64$x higher in Hybrid than ODN. Among them, the number of correctly detected objects is $1.596$x more for Hybrid than ODN. The ratio of detecting the same object overlapping is $8.82$\% for Hybrid and $3.12$\% for ODN. The percentage of falsely detecting objects is $6.77$\% for ODN and $3.42$\% for Hybrid, which is lower in Hybrid. In the total number of detected objects, compared to the number of objects excluding overlapping objects, the number of objects accurately detected is $93.22$\% for ODN and $96.24$\% for Hybrid, showing the superiority in the accuracy of $3.02$\% for Hybrid.\\

\begin{table}[t!]
\footnotesize
    \centering
        \caption{The inference time of three models with various settings.}
    \label{tab:inf_time}
    \begin{tabular}{c|c|c|c}
    \toprule[1pt]
    \textbf{Settings} & \textbf{Hybrid} & \textbf{Real-time ODN} & \textbf{Complex ODN}\\\midrule
    \multicolumn{1}{c|}{GPU}        & 83\,ms & 55\,ms & 102\,ms \\
    \multicolumn{1}{c|}{CPU}        &133\,ms & 67\,ms & 562\,ms \\
    \bottomrule[1pt]
    \end{tabular}
\end{table}
\begin{table}[t!]
\footnotesize
    \centering
        \caption{The performance comparison between two models for the number of the detected object.} 
    \label{tab:my_label}
    \begin{tabular}{c|c|c}
    \toprule[1pt]
    \textbf{Metric} & \textbf{ODN} & \textbf{Hybrid (Ours)}\\\midrule
    \multicolumn{1}{l|}{Total object} & 4,603 & 7,562\\
    \multicolumn{1}{l|}{Correctly Detected}  & 4,157 & 6,636\\
    \multicolumn{1}{l|}{Falsely Detected}   & 312 & 259\\
    \multicolumn{1}{l|}{Overlapped Detected}  & 144 & 667\\ 
     \multicolumn{1}{l|}{True Positive Rate(\%)}  & 93.22 & 96.24\\ 
    \bottomrule[1pt]
    \end{tabular}
\end{table}
\subsection{Implementation of Model Selection}\label{sec:3-2}
\begin{figure}[t]
\centering
\setlength{\tabcolsep}{2pt}
\renewcommand{\arraystretch}{0.2}
\begin{tabular}{cc}
\multicolumn{2}{c}{\includegraphics[page=1, width=0.5\textwidth]{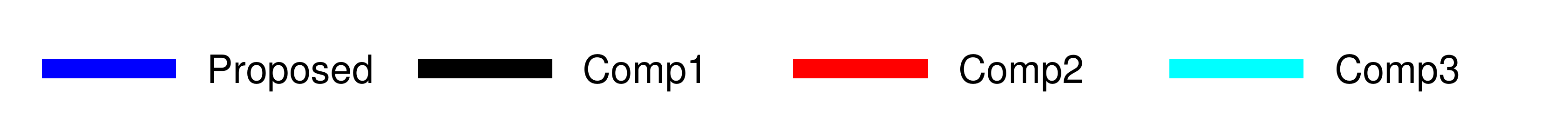}}\\
\includegraphics[page=1,width=0.24\textwidth]{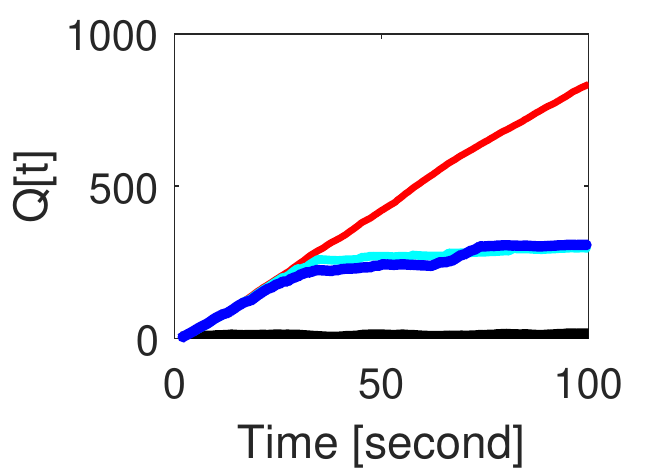} &  \includegraphics[page=1,width=0.24\textwidth]{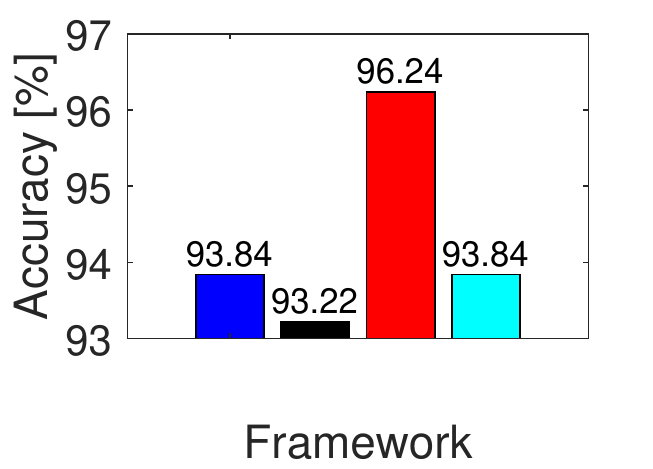}    
\\
\\
\small (a) Queue-backlog & \small (b) Accuracy on average
\end{tabular}
\caption{The results of \textit{model selection} }
\label{fig:performance2}
\end{figure} 

\begin{table}[t!]
\footnotesize
    \centering
        \caption{The configuration of the Comp3 framework}
    \label{tab:comp3}
    \begin{tabular}{l|r}
    \toprule[1pt]
    \textbf{Element} & \textbf{Detail}\\\midrule[1pt]
    \multicolumn{1}{l|}{\multirow{2}{*}{State}} & $s_t=\{Q[t], a[t-1], b(\alpha[t-1]), b(\alpha[t]), $\\
    &   $P(\alpha[t]),~w_1,~w_2,~w_{fps},~w_p,~V\}$\\\midrule
    \multicolumn{1}{l|}{Action}  & $a_t\in\mathcal{A}\equiv\{H,T\}$\\\midrule
    \multicolumn{1}{l|}{Reward}   & $r_t = V\cdot P(\alpha[t])+Q[t]b(\alpha[t])$ \\\midrule
    \multicolumn{1}{l|}{Optimization method}  & REINFORCE \cite{sutton1999policy} \\ \midrule
    \multicolumn{1}{l|}{Optimizer}  & Adam optimizer \\ \midrule
    \multicolumn{1}{l|}{Learning rate}  & 0.0002 \\ \midrule
        & 3 fully connected layers with ReLU function\\
       \multicolumn{1}{l|}{\multirow{1}{*}{Neural Network}} & \multicolumn{1}{l}{~~- 1st layer: $10\times 128$ + ReLU} \\
       \multicolumn{1}{l|}{\multirow{1}{*}{Architecture}} & \multicolumn{1}{l}{~~- 2nd layer: $128\times 128$ + ReLU}\\
        & \multicolumn{1}{l}{~~- 3rd layer: $128\times 2$ + ReLU}\\\bottomrule[1pt]
    \end{tabular}
\end{table}
We investigate the performance to verify the Lyapunov optimization framework, which chooses between ODN and the Hybrid in a finite time (\textit{e.g.}, 100 seconds) depending on the amount of queue-backlog (\textit{i.e.}, delay). For comparison with the proposed framework, we adopt three comparison frameworks, \textit{i.e.}, the policy with selecting only ODN, or only the Hybrid and the deep reinforcement learning (DRL)-based policy, which are denoted as `Comp1', `Comp2', and `Comp3', respectively. To configure Comp3, we design the state, action, reward, optimization methods, and neural network architecture depicted in Table~\ref{tab:comp3}. Note that Comp3 is the state-of-the-art stochastic controlling method.

\BfPara{Numerical result of model selection}
Fig.~\ref{fig:performance2} shows the result of the network in which Hybrid and ODN observe the queue and thereby select a model.
In the cases of model selection based on the proposed model selection and Comp3 on the mobile platform, it is confirmed that neither transient stability nor queue overflow occurs. In contrast, transient stability occurs in Comp1. In addition, queue overflow occurs in Comp2. Thus, our proposed framework and Comp3 outperform Comp1 and Comp2, corresponding to the queue stability (see Fig.~\ref{fig:performance2}(a)) and average accuracy (see Fig.~\ref{fig:performance2}(b)).

\BfPara{Computing cost for model selection}
Comp3 consumes $35,582$\,FLOPS for every model selection regarding computing cost. Furthermore, Comp3 requires additional computing costs for training the policy. However,  the computing cost of every model selection requires only $12$\,FLOPS for the proposed framework, whereas no computing cost is necessary for Comp1, Comp2. The proposed framework is most suitable for model selection for the real-time service and among all frameworks. Consequently, we corroborate that the proposed model selection framework guarantees time-average performance maximization and queue stability, which requires very low-computational cost for model selection.
 
\section{Conclusions and Future Work}\label{sec:4}
In this paper, we confirm the improved performance of ODN in the road driving environment by utilizing optical flow estimation via OFEN and flow map processing. 
The flow map processing algorithm can be applied to all ODNs that utilize bounding box detection. A Lyapunov optimization-based model selection framework is also constructed to select a model under autonomous driving environments' constraints automatically. In addition, it is known to be time-average optimal subject to system/queue stability. As a result of observing the queue-backlog, it is confirmed that the time-average performance maximization can be achieved, as theoretically expected. As future research directions, real-world experimental study results in an actual driving environment are meant to verify the novelty of this algorithm in practice.

\bibliographystyle{IEEEtran}
\balance
\bibliography{ref}

 \end{document}